\documentclass{bmvc2k}

\usepackage{multirow}
\usepackage{adjustbox}
\usepackage{colortbl}

\usepackage{graphicx} % Required for including images
\usepackage{wrapfig}

\title{Advancing Compressed Video Action Recognition through Progressive Knowledge Distillation}

\addauthor{Efstathia Soufleri}{esoufler@purdue.edu}{1}
\addauthor{Deepak Ravikumar}{dravikum@purdue.edu}{1}
\addauthor{Kaushik Roy}{kaushik@purdue.edu}{1}

\addinstitution{
Electrical and Computer Engineering,\\ Purdue University,\\ 
West Lafayette, USA
}
\runninghead{Efstathia Soufleri, Deepak Ravikumar, Kaushik Roy}{PKD and WISE}

\begin{document}

\maketitle

\begin{abstract}
Compressed video action recognition classifies video samples by leveraging the different modalities in compressed videos, namely motion vectors, residuals, and intra-frames. For this purpose, three neural networks are deployed, each dedicated to processing one modality. Our observations indicate that the network processing intra-frames tend to converge to a flatter minimum than the network processing residuals, which in turn converges to a flatter minimum than the motion vector network. This hierarchy in convergence motivates our strategy for knowledge transfer among modalities to achieve flatter minima, which are generally associated with better generalization. With this insight, we propose Progressive Knowledge Distillation (PKD), a technique that incrementally transfers knowledge across the modalities. This method involves attaching early exits (Internal Classifiers - ICs) to the three networks. PKD distills knowledge starting from the motion vector network, followed by the residual, and finally, the intra-frame network, sequentially improving IC accuracy. Further, we propose the Weighted Inference with Scaled Ensemble (WISE), which combines outputs from the ICs using learned weights, boosting accuracy during inference. Our experiments demonstrate the effectiveness of training the ICs with PKD compared to standard cross-entropy-based training, showing IC accuracy improvements of up to 5.87\% and 11.42\% on the UCF-101 and HMDB-51 datasets, respectively. Additionally, WISE improves accuracy by up to 4.28\% and 9.30\% on UCF-101 and HMDB-51, respectively.\footnote{The PyTorch code is available at \url{https://github.com/Efstathia-Soufleri/PKD}.}
\end{abstract}

\section{Introduction}

\begin{figure}[t]
    \centering
    \begin{minipage}[t]{0.38\textwidth}
        \centering
        \includegraphics[width=\textwidth]{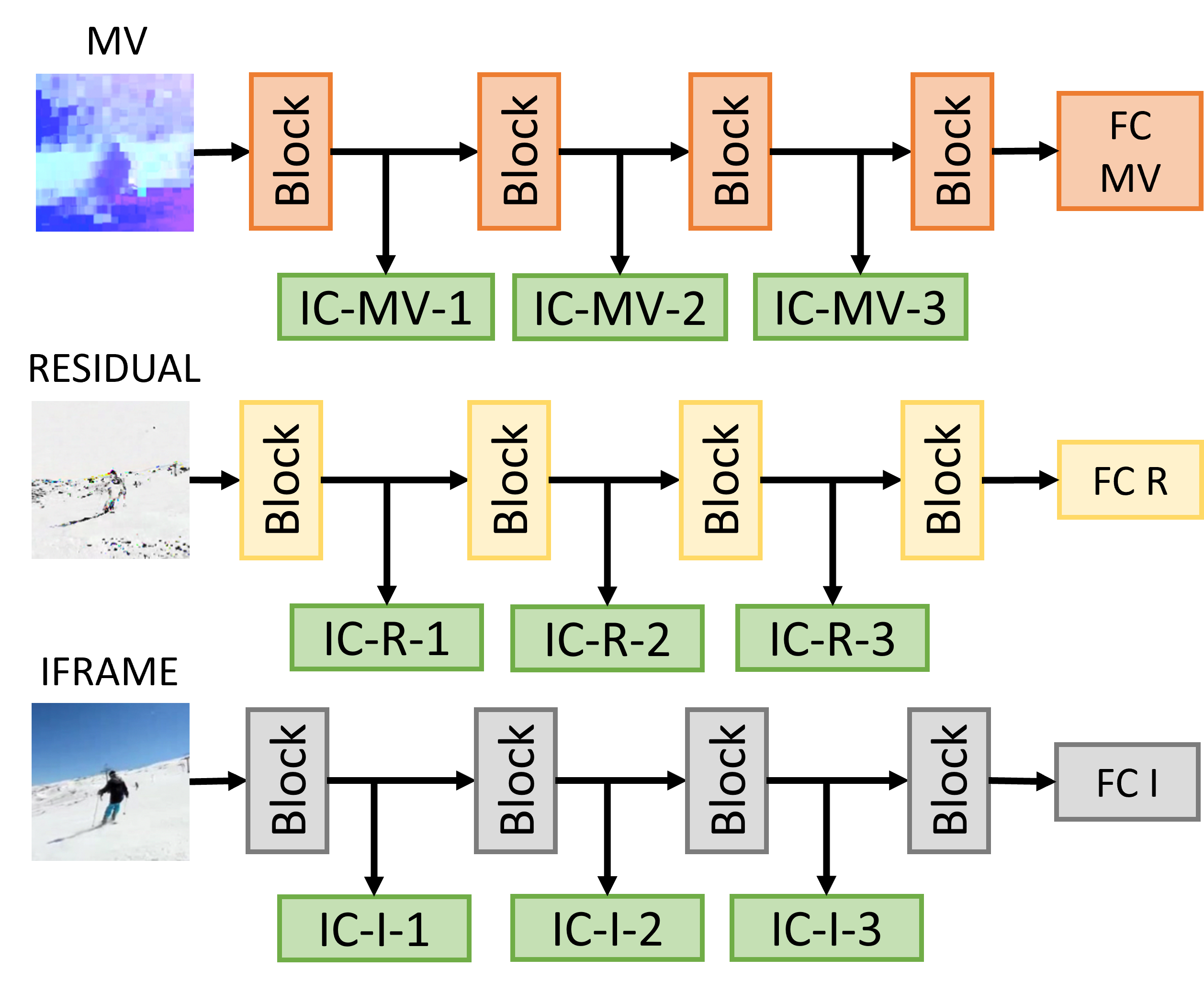} %images/intro-ic-v2.png
        \caption{Backbone networks with ICs for each video modality.}
        \label{fig:architecture}
        \vspace{-2mm}
    \end{minipage}\hfill
    \begin{minipage}[t]{0.58\textwidth}
        \centering
        \includegraphics[width=\textwidth]{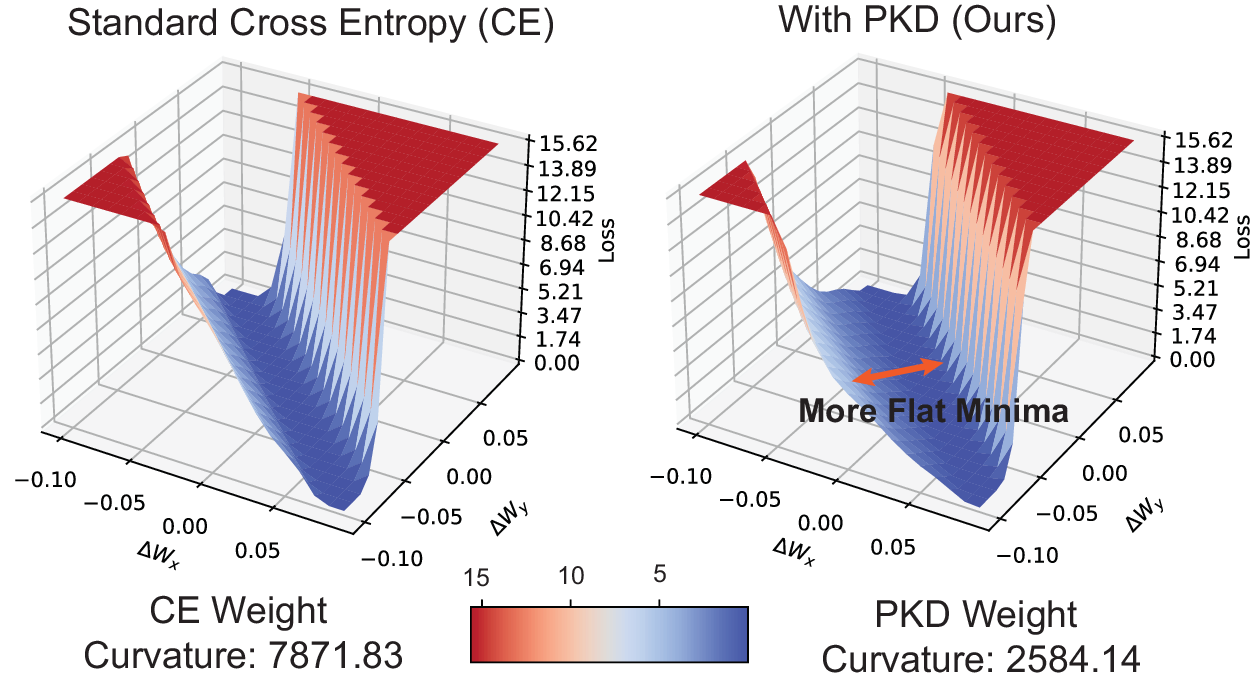} % Second image
        \caption{Visualizing flatter minima resulting from PKD and the associated loss curvature.}
        \label{fig:flatter_minima}
        \vspace{-2mm}
    \end{minipage}
    \vspace{-4mm}
\end{figure}

In typical video classification, the camera captures video data, which is then compressed using video compression algorithms (such as MPEG-4, H.264, and HEVC) \cite{richardson2011h,duan2020video,wiegand2003overview}, transmitted, and subsequently decompressed by the receiver for classification. For such classification, the bottlenecks are the time-consuming compression and decompression steps. To alleviate the above bottleneck, video classification directly from compressed videos using motion vectors (MV), residuals (R), and intra-frames (I-frames) has gained significant attention in recent years \cite{wu2018compressed,wang2021team,liu2023learning,battash2020mimic,shou2019dmc,dos2019cv}. For example, CoViAR \cite{wu2018compressed} deploys three neural networks, one for processing each compressed video modality (MV, R, I-frames). We observe that networks trained on I-frames converge to a flatter minima, compared to networks trained on R, which in turn are flatter than those trained on MV. Models with flatter minima have been shown to generalize better \cite{hochreiter1994simplifying, hochreiter1997flat}. In this paper, we leverage these insights to build a more efficient compressed video classification framework.

Firstly, to improve training, we propose progressive knowledge distillation (PKD). The goal of PKD is to progressively distill knowledge from the models with flatter minima to models with less flat minima to improve their performance. Similar to previous works \cite{wu2018compressed,wang2021team}, we deploy three neural networks to process the MV, R, and I-frame modalities: MV backbone network, R backbone network, and I-frame backbone network. We propose distilling each modality progressively using more flat networks. That is distill in a sequence using MV, R and I-frame backbones as teachers progressively. 

However, this approach presents a cyclic dependency issue. Specifically, training the MV backbone requires all backbones to be trained, and similarly for the R and I-frame backbones. To overcome this problem, we propose using early exit classifiers. Specifically, we train a  backbone network using standard cross-entropy (CE) loss for each modality. After the backbone networks have converged, their parameters are frozen. To the frozen backbone, we attach Internal Classifiers (ICs) for early exit, as shown in Figure \ref{fig:architecture}. These ICs are trained using PKD. The teacher models are the final classifiers (FCs) of the backbone networks of the MV, R, and I-frame. The students are the ICs. Our proposal, PKD trains the ICs in three steps: (1) distill knowledge from the FC of the MV backbone network to all the ICs, (2) distill knowledge from the FC of the R backbone network to all the ICs, and (3) distill knowledge from the FC of the I-frame backbone network to all the ICs. During knowledge distillation, the parameters of the ICs are updated while the parameters of the backbone networks remain frozen. We show that PKD for IC training results in a flatter minima compared to CE (Figure \ref{fig:flatter_minima} and better performance and thus, \emph{improved efficiency}.

While PKD tackles training, to improve inference efficiency, we propose Weighted Inference with a Scaled Ensemble (WISE). Consider the $l^{th}$ IC, WISE combines the prediction of the $l^{th}$ IC with predictions from the previous $l-1$ ICs. Each of the $l$ predictions is multiplied by a corresponding scaling factor and aggregated to output a prediction at the $l^{th}$ exit. If the confidence of this prediction is greater than a predefined threshold value, then classification for this video sample terminates at this exit. Otherwise, the classification continues. WISE formulates an optimization problem to determine the scaling factors. Our experimental evaluation demonstrates an increase in classification accuracy by up to 4.28\% on UCF-101 and 9.30\% on HMDB-51 when using WISE.

In summary, the paper's main contributions are: (1) We propose PKD, a novel IC training methodology to transfer knowledge progressively from the FC of the MV, R, and I-frame backbone networks to the ICs, resulting in \emph{improved efficiency}. (2) We propose WISE, an efficient inference methodology. WISE leverages cross-modality predictions of previous ICs to improve inference \emph{efficiency}. (3) Experimental results of our proposal on UCF-101 and HMDB-51 datasets show up to $\approx 11\%$ IC accuracy improvement with PKD compared to CE. Moreover, we observe up to $\approx$ 9\% accuracy improvement when using WISE.

\vspace{-3mm}
\section{Related Work}
\label{Related Work}

Video compression algorithms, including established standards like MPEG-4 \cite{ebrahimi2000mpeg}, H.264 \cite{wiegand2003overview}, and HEVC \cite{sullivan2012overview}, exploit the recurring similarity between successive frames in a video sequence. These modern codecs typically partition videos into intra-frames (I-frames) and predictive frames (P-frames). I-frames are essentially regular images, preserving spatial information, and are compressed similarly to regular images. P-frames encode temporal information by capturing the ``changes'' between the frames over time. P-frames consist of MV and R. MVs capture the coarse block-wise movement between frames, while the residuals capture the pixel-wise differences between frames. MV, R, and I-frames are referred as compressed video modalities.

Prior works can be divided into three categories: 1) works that use both compressed and raw (uncompressed) videos during training to enhance accuracy, 2) works that use only compressed videos but have high computational cost, and 3) works that use only compressed videos and focus on reducing the computational cost. 

Works that fall in the first category achieve high classification accuracy but come at a significant computational cost. The authors of compressed modality distillation \cite{liu2023learning} propose using multiple networks and mechanisms (such as ResNet50 networks, Bi-ConvLSTM, self-attention mechanisms, and temporal graphs) to learn and capture the relationship between the raw and the compressed video domains to boost classification accuracy. The authors of MFCD-Net \cite{battash2020mimic} propose a network architecture suitable for processing both the raw and the compressed video. For these works, the benefit of using compressed videos diminishes as access to the raw video during training is required.

CoViAR \cite{wu2018compressed} was one of the earliest compressed video action recognition works in the second category. CoViAR uses the following backbone networks: a ResNet152 to classify the I-frames, a ResNet18 to classify the R, and a ResNet18 to classify the MV. Then, the predictions from these three backbone networks are combined by averaging them. While CoViAR achieves satisfactory classification accuracy, it comes at a significant computational cost. Moreover, TEAM-Net \cite{wang2021team} is based on CoViAR, uses similar backbone networks (ResNet50 instead of ResNet152 for the I-frame), and proposes a new module that captures the temporal and spatial relationship of the compressed modalities. CV-C3D \cite{dos2019cv} extends the 3D convolutional neural networks in the compressed domain. DMC-Net \cite{shou2019dmc} follows an approach similar to CoViAR, but uses an additional network trained using Optical Flow (OF).

Finally, we discuss works that focus on reducing the computational cost. The authors of multi-teacher knowledge distillation \cite{wu2019multi} use the same backbone networks as CoViAR but replace the ResNet152 with a ResNet18. Their backbone networks are trained using multi-teacher knowledge distillation. However, distilling knowledge from ResNet152 to ResNet18 leads to a noticeable drop in network accuracy. Based on the independent sub-network training \cite{havasi2020training}, the authors of MIMO \cite{terao2023efficient} propose a single network that has multiple-inputs-multiple-outputs (MIMO). The MIMO network concatenates the MV, R, and I-frames and gives them as input to the network to perform a single forward pass. This methodology reduces computational cost but imposes accuracy limitations constrained by the network's capacity. Further, the coupled nature of the input imposes the use of an equal number of MV, R, and I-frame as input. This constraint hinders real-time applications, where the entire video stream is not available in advance. Our work falls in this third category and scales between classification accuracy and computational cost.

\vspace{-2mm}
\section{Methodology}
\label{Methodology}
 
\subsection{Progressive Knowledge Distillation (PKD)}
\label{Prog_KD}

\begin{figure*}[t]
\centering
\includegraphics[width=0.9\textwidth]{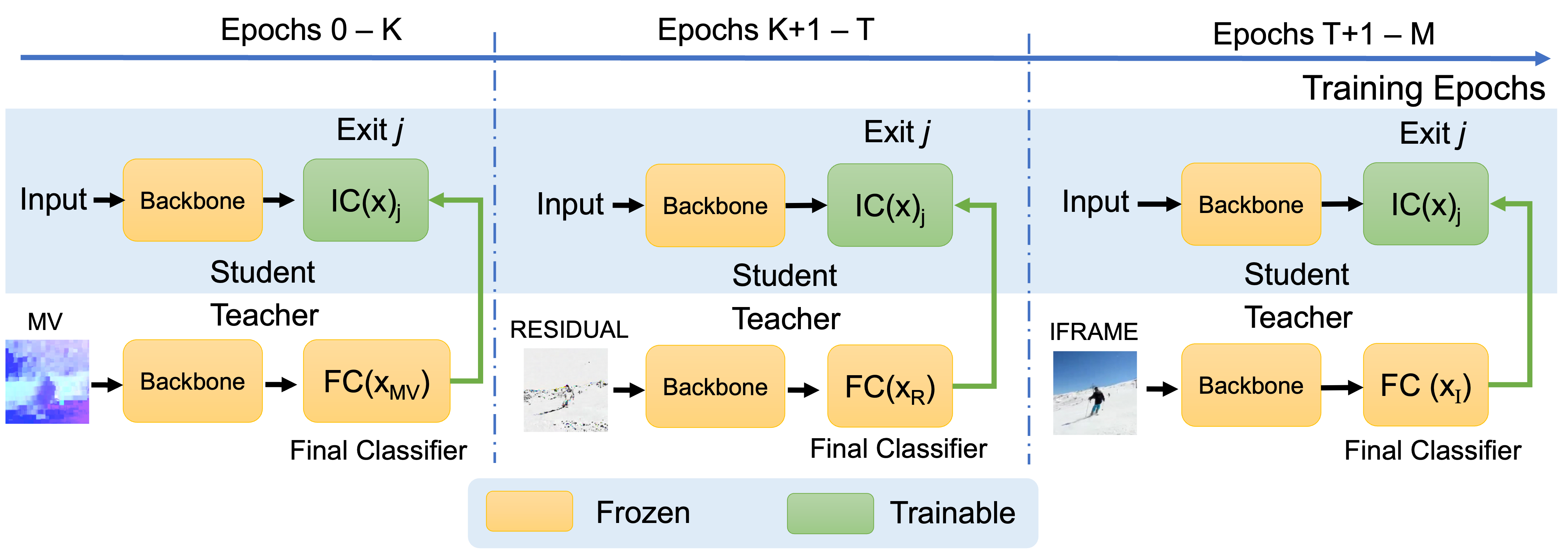}
\caption{PKD IC training Overview: 1) for epoch 0 till K, we perform KD between the IC and the final classifier (FC) of the MV backbone network, 2) for epoch K+1 till T, we perform KD between the IC block and FC of the R backbone network, 3) for epoch T+1 till M, we perform KD between the IC block and the FC of the I-frame backbone network. Note that, the parameters of the backbone networks for the MV, R, and I-frame (illustrated in yellow) are not updated during PKD. The ICs (illustrated in green) are trained independently.}
\label{prog_kd_figure}
\vspace{-4mm}
\end{figure*}

PKD is based on the observation that neural networks trained on I-frames converge to a flatter minima compared to models trained on R when converge to flatter minima that models trained on MV. Flatter minima is known to generalize better \cite{hochreiter1994simplifying, hochreiter1997flat}. To leverage this, we propose PKD which progressively distills knowledge from the compressed video modalities.

The training process consists of two phases. In the first phase, the MV, R, and I-frame backbone networks are trained. The input to the MV backbone network is the motion vectors. Similarly, the R and I-frame backbone networks classify the R and I-frame inputs, respectively. For training the backbone networks, we minimize the CE loss between the predictions and the actual labels. This loss is back-propagated from the FC of the backbone network to the first layer, updating the parameters of the backbone network. Each MV, R and I-frame backbone network is trained independently from the other. 

Upon convergence of the backbone network its weight parameters are frozen in preparation for the next phase. In the next phase, the ICs are attached to the trained MV, R, and I-frame backbone networks. The ICs are trained with our proposed PKD methodology. The trained FCs from each backbone network are used as teachers to perform Knowledge Distillation (KD) on the ICs. PKD is based on the idea that transferring knowledge across the compressed video modalities (i.e. MV, R, I-frame) yields more efficient training with flatter minima, compared to using CE or KD from a single classifier. This progressive knowledge transfer consists of three steps, as illustrated in Figure \ref{prog_kd_figure}. (1) MV-KD: KD between the FC of the MV backbone network and the ICs. (2) R-KD: KD between the FC of the R backbone network and the ICs. (3) I-frame-KD: KD between the FC of the I-frame backbone network and the ICs. 

Note that during PKD, each IC is trained independently, with each KD loss backpropagating only to the layers of the respective IC. Further, in the description provided we assume that the backbone used is the CoViAR framework, but the same can be applied to other compressed video classification backbones as shown in the experiments section.

\subsection{Weighted Inference with Scaled Ensemble (WISE)}
\label{WISE}

\begin{figure*}[t]
\centering
\includegraphics[width=0.9\textwidth]{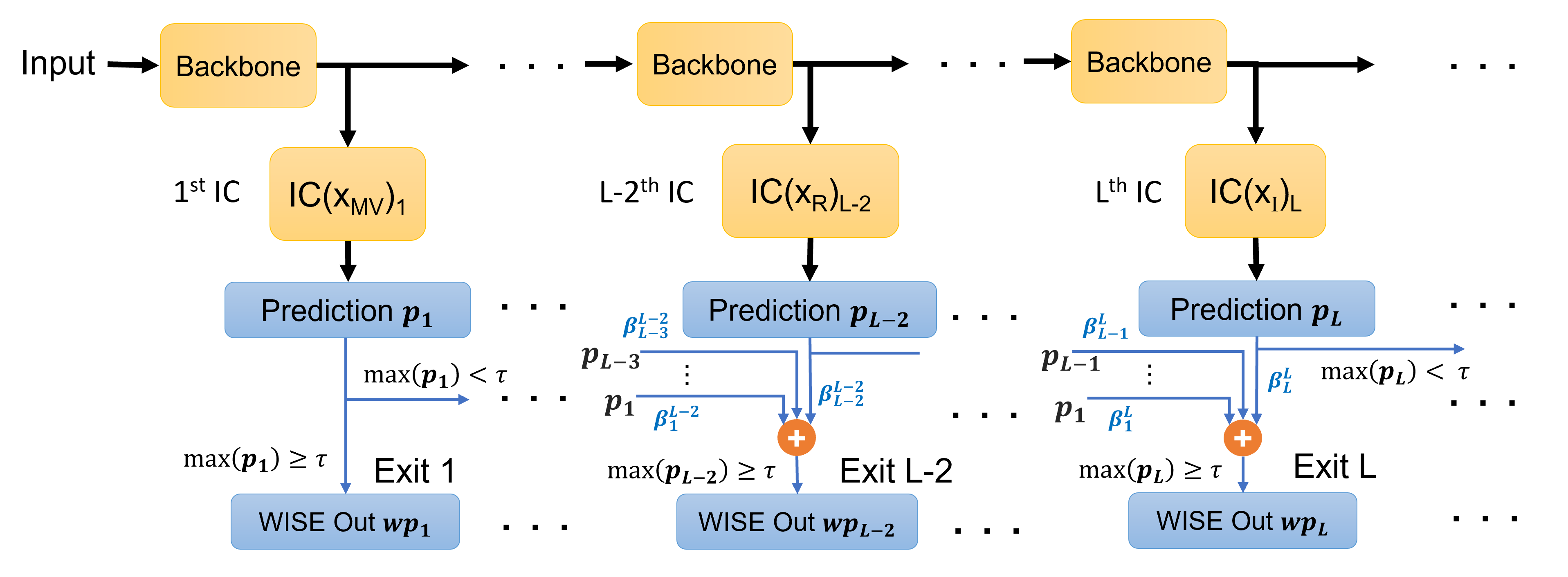}
\caption{WISE Inference Overview: the video sample is evaluated sequentially. The exits might be from different compressed video modality backbones. The previous IC predictions are combined with scaling factors $\beta$ into an ensemble. If the confidence of the prediction exceeds a certain threshold $\tau$, classification terminates. Otherwise, the next IC is evaluated.}
\label{wise_figure}
\vspace{-4mm}
\end{figure*}

During inference, we deploy the trained ICs and each IC is a potential early exit. An IC is an exit point if the confidence of the prediction exceeds a certain threshold $\tau$ (see Figure \ref{wise_figure}). In this case, the classification process terminates at that exit point; otherwise, it proceeds to the next IC. To that effect, we propose Weighted Inference with Scaled Ensemble (WISE). WISE combines the predictions from previous early exits (i.e. ICs where confidence was less than $\tau$). This is achieved by multiplying each IC prediction with scaling factors, followed by aggregation.

For example, let's consider that we are at the $L^{th}$ exit IC. At the $L^{th}$ exit we would have had information from the previous $L-1$ exits. With WISE, we combine the predictions from these ICs and we propose a simple linear combination, where the predictions from the previous ICs are multiplied with some scaling factors $\beta$ and aggregated to an ensemble prediction. Figure \ref{wise_figure} illustrates WISE. To find the optimal scaling factors for each IC prediction, we formulate an optimization problem. This optimization problem aims to minimize the CE loss between the ensemble predictions and the actual labels over the training set. The optimization problem is defined as:

\begin{align}
    \min_{\beta^L} & \quad -\cfrac{1}{N}\sum_{i=1}^{N} t_i \log \left(\sum_{j=1}^{L}\beta^L_j \cfrac{exp(IC_{i}^{j})}{\sum_{m=1}^{m=k}exp(IC_{m}^{j})} \right) 
\end{align}
where $N$ is the number of video samples in the training set, $t_i$ is the actual label for the $i^{th}$ video sample, and $L$ is the $L^{th}$ IC being optimized. Note that $IC^j \in R^{k}$ is an $k$ dimensional logit from the $j^{th}$ IC for a $k$ class classification problem. $IC_{m}^{j}$  is the $m^{th}$ element of $IC^j$ and $\beta^L = [\beta^L_1, \cdots, \beta^L_L]$ are the scaling factors to combine the IC predictions from the various ICs at the $L^{th}$ exit. Thus, the optimization problem determines the scaling factors $\beta$, which combines predictions from various ICs from different modalities. The networks are evaluating sequentially starting from R, then continuing to MV and finally I-frame network. An IC attached to the I-frame backbone has scaling factors $\beta$ that combine the predictions from previous ICs attached to the MV and R backbone networks as well as the previous ICs of the I-frame backbone network.

\vspace{-2mm}

\section{Experimental Evaluation}
\label{Experimental Evaluation}

\subsection{Datasets and Implementation Details}
\label{sec:setup}

\textbf{Datasets.} We evaluated our method on two action recognition datasets: UCF-101 \cite{soomro2012ucf101} and HMDB-51 \cite{kuehne2011hmdb}. UCF-101 contains 13,320 videos from 101 action categories. HMDB-51 contains 6,766 videos from 51 action categories. Each video in both datasets is annotated with one action label. Each dataset has 3 training and testing splits for training and evaluation. We report the average accuracy of the three testing splits over three different seeds for all of our results (i.e. 3 seeds for each split thus, 9 data points in total for one entry). 

\noindent \textbf{Architecture and Backbone.} We used two types of backbone networks: 1) CoViAR \cite{wu2018compressed}, and 2) TEAM-Net \cite{wang2021team}. CoViAR consists of a ResNet18 \cite{he2016deep} with temporal shift modules (TSMs) \cite{lin2019tsm} for the MV, a ResNet18 with TSM modules for the R, and a ResNet50 \cite{he2016deep} with TSMs for the I-frame network. Note that we use a ResNet50 instead of ResNet152 for the CoViAR backbone. The use of TSMs is a standard tool for action recognition tasks and facilitates the adaptation of 2D convolutional networks for video processing without increasing the parameters and floating-point operations (FLOPs). ICs are attached after each ResNet block. TEAM-Net is similar to CoViAR (uses a ResNet18 for MV, a ResNet18 for R, and a ResNet50 for I-frame), but after each block, it inserts an additional block (TEAM block) that concatenates the features of the three compressed modalities, passes them through a convolutional layer, and then separates them again using three fully connected layers. We attach the ICs after each TEAM block. We focused on the CoViAR backbone for all the experiments, which we found to be more efficient when scaling. However, we also present the results for the TEAM-Net backbone. The datasets and architectures used are standard practice \cite{wu2018compressed, wang2021team,battash2020mimic,dos2019cv}. Details about training and inference are provided in the Appendix.
\vspace{-2mm}
\subsection{Evaluating PKD}
\label{Evaluating PKD}

\textbf{Experiment.}
We train the ICs attached to the MV, R, and I-frame backbone networks using two approaches: 1) PKD, and 2) CE. We evaluate the accuracy of each IC on the test set. We use 3 frames for each compressed video modality (I-frame, R, MV) for CoViAR and 8 frames for TEAM-Net. We present the accuracy of the ICs on UCF-101 and HMDB-51 in Table \ref{table_combined_results}. Since the backbone networks have 4 blocks, attaching an IC after each block results in 3 ICs per modality (i.e. 3 for each MV, R, and I-frame). Note that the $4^{th}$ IC is the final classifier and is not trained using PKD or CE, but is trained in the pre-training phase and thus is excluded from this comparison.

\noindent \textbf{Results.}
The classification accuracy of the ICs are presented in Table \ref{table_combined_results}. IC1, IC2, and IC3 correspond to the ICs attached after the first block, second block, and third block respectively for classifying all the video samples. To highlight the improvement of the proposed PKD over CE we also present a row with the difference between CE and PKD. The results in Table \ref{table_combined_results} demonstrate the superior accuracy of ICs trained with PKD using CoViAR as the backbone framework, showing an increase in accuracy of up to 1.77\%, 2.72\%, and 5.87\%, for MV, R, and I-frame, respectively on UCF-101 and 3.50\%, 7.87\% and 11.42\% on HMDB-51 when compared to the ICs trained with CE. When using TEAM-Net as the backbone framework, we observe an increase in IC accuracy of up to 1.24\%, 2.32\%, and 2.79\%, for MV, R, and I-frame, respectively on UCF-101 and 0.33\%, 1.08\% and 1.65\% on HMDB-51, when using PKD compared to training the ICs with CE loss. Note, that this improvement sources from the fact that the ICs trained with PKD result in flatter minima compares to the ICs trained with CE, and thus generalize better. Figure \ref{fig:flatter_minima} visualizes the landscape for IC2 of I-frame. The remaining visualizations are in the Appendix.

{\renewcommand*{\arraystretch}{1.1}
\begin{table*}[t]
\centering
\begin{adjustbox}{width=\textwidth}
\begin{tabular}{|c|c|ccc|ccc|ccc|ccc|c|}
\hline
 &  & \multicolumn{12}{c|}{Backbones}  \\ \cline{3-14} 
\multirow{3}{*}{Dataset} & \multirow{3}{*}{Method} & \multicolumn{9}{c|}{CoViAR} & \multicolumn{3}{c|}{TEAM-Net} \\ \cline{3-14} 
& &  \multicolumn{3}{c|}{MV} & \multicolumn{3}{c|}{Residual}  & \multicolumn{3}{c|}{I-Frame} & \multirow{2}{*}{IC1}  &  \multirow{2}{*}{IC2} &  \multirow{2}{*}{IC3}\\ 
& & IC1 & IC2 & IC3 & IC1 & IC2 & IC3 & IC1 & IC2 & IC3 &  &  &  \\ \hline \hline
\multirow{3}{*}{UCF-101} & CE & 34.39 & 47.29 & 57.43 & 46.45 & 60.57 & 72.63 & 48.40 & 57.61 & 74.11 & 8.86 & 28.93 & 47.41 \\
& Prog KD (Ours) & 35.10 & 48.67 & 59.20 & 47.17 & 62.23 & 75.34 & 49.11 & 60.72 & 79.98 & 10.10 & 31.25 & 50.20 \\
& Diff. Prog KD $-$ CE & \textbf{0.71} & \textbf{1.38} & \textbf{1.77} & \textbf{0.72} & \textbf{1.66} & \textbf{2.72} & \textbf{0.71} & \textbf{3.11} & \textbf{5.87} & \textbf{1.24} & \textbf{2.32} & \textbf{2.79} \\ \hline \hline
\multirow{3}{*}{HMDB-51} & CE & 19.92 & 26.24 & 34.58 & 21.78 & 31.13 & 40.19 & 23.55 & 27.09 & 42.42 & 5.38 & 18.12 & 24.40 \\
& Prog KD (Ours) & 20.10 & 27.43 & 38.08 & 24.77 & 35.27 & 48.07 & 27.32 & 35.15 & 53.84 & 5.71 & 19.21 & 26.04 \\
& Diff. Prog KD $-$ CE & \textbf{0.18} & \textbf{1.20} & \textbf{3.50} & \textbf{2.99} & \textbf{4.14} & \textbf{7.87} & \textbf{3.77} & \textbf{8.06} & \textbf{11.42} & \textbf{0.33} & \textbf{1.08} & \textbf{1.65} \\ \hline
\end{tabular}
\end{adjustbox}
\caption{Comparing the results of the proposed PKD method for IC1, IC2, and IC3 with Cross Entropy (CE) using MV, R, and I-frames. The reported accuracy corresponds to the average of the three testing splits of on UCF-101 and HMDB-51 datasets for results over three different seeds and two backbone frameworks CoViAR and TEAM-Net.}
\label{table_combined_results}
\vspace{-3mm}
\end{table*}}

\subsection{Evaluating WISE}
\label{Evaluating WISE}

We experimentally analyze the effectiveness of the proposed WISE methodology during inference. WISE has two key features: 1) the use of previous IC predictions (via lateral connections) available at the exit point to capture cross-modality information during inference, and 2) the efficient combination of these predictions using scaling factors. As a comparison, we evaluate WISE versus three alternative scenarios: 1) a baseline scenario not utilizing previous IC predictions (no lateral connections), 2) using previous ICs with uniform scaling factors, and 3) utilizing scaling factors obtained from WISE.  Note, that lateral connections at the $L^{th}$ exit point means the use of predictions from the previous $L-1$ ICs.

\noindent \textbf{Experiment.}
For this experiment, we evaluate the three methods (no lateral connections, uniform scaling, WISE) and select a threshold for the exit points such that all the methods have iso-computational cost. Note when using lateral connections with uniform scaling, $\beta$ is set to $1$, that is all the previous predictions have equal weighting on the output. The accuracy of the three methods on UCF-101 and HMDB-51 datasets are shown in Table \ref{table_wise}. 

\begin{table}[t]%[b!]
\centering
\begin{tabular}{|c|cccc|}
\hline
                                               & \multicolumn{4}{c|}{Backbones}                                                                       \\ \hline
                                               & \multicolumn{2}{c|}{COVIAR}                                 & \multicolumn{2}{c|}{TEAM-Net}          \\ \hline
Method                                         & \multicolumn{1}{c|}{UCF-101} & \multicolumn{1}{c|}{HMDB-51} & \multicolumn{1}{c|}{UCF-101} & HMDB-51 \\ \hline 
\multicolumn{1}{|c|}{No Lateral   Connections} & \multicolumn{1}{c|}{88.59\%} & \multicolumn{1}{c|}{58.91\%} & \multicolumn{1}{c|}{93.24\%} & 65.94\% \\ \hline
\multicolumn{1}{|c|}{Uniform   Scaling (Ours)} & \multicolumn{1}{c|}{92.08\%} & \multicolumn{1}{c|}{67.50\%} & \multicolumn{1}{c|}{89.49\%} & 61.89\% \\ \hline
\multicolumn{1}{|c|}{WISE (Ours)}              & \multicolumn{1}{c|}{92.87\%} & \multicolumn{1}{c|}{68.21\%} & \multicolumn{1}{c|}{93.29\%} & 66.29\% \\ \hline
\end{tabular}
\caption{Result Summary when using: 1) no lateral connections between the ICs, 2) uniform scaling factors and lateral connections between the ICs, 3) WISE. CoViAR and TEAM-Net the used backbone networks. The reported accuracy is the average over the \emph{three testing splits} and \emph{three different seeds} (9 datapoints per entry) for UCF-101 and HMDB-51.}
\label{table_wise}
\end{table}

\noindent \textbf{Results.}
The accuracy of each approach is summarized in Table \ref{table_wise}. The results demonstrate the comparative accuracy of the baseline (no lateral connections), uniform scaling, and WISE scenarios. WISE outperforms the baseline and uniform scaling by $\approx$ 4.28\% and 0.8\% respectively when using CoViAR as the backbone, and by $\approx$ 0.1\% and 4\% using TEAM-Net as the backbone on UCF-101. Furthermore, WISE outperforms the baseline and uniform scaling by $\approx$ 9\% and 1\% respectively when using CoViAR as the backbone, and by 0.4\% and 4\% using TEAM-Net as the backbone network on HMDB-51.

% \vspace{-2mm}
{\renewcommand*{\arraystretch}{1.1}
\begin{table*}[t]
\centering
\begin{adjustbox}{width=0.95\textwidth}
\begin{tabular}{|c|c|ccc|ccc|ccc|ccc|}
\hline
 &  & \multicolumn{12}{c|}{Backbones}  \\ \cline{3-14} 
\multirow{3}{*}{Dataset} & \multirow{3}{*}{Method} & \multicolumn{9}{c|}{CoViAR} & \multicolumn{3}{c|}{TEAM-Net} \\ \cline{3-14} 
& &  \multicolumn{3}{c|}{MV} & \multicolumn{3}{c|}{Residual}  & \multicolumn{3}{c|}{I-Frame} & \multirow{2}{*}{IC1}  &  \multirow{2}{*}{IC2} &  \multirow{2}{*}{IC3}\\ 
& & IC1 & IC2 & IC3 & IC1 & IC2 & IC3 & IC1 & IC2 & IC3 &  &  &  \\ \hline \hline
\multirow{3}{*}{UCF-101 } & I-frame KD               & 34.51 & 47.19 & 57.38 & 46.70 & 60.92 & 73.25 & 48.93 & 59.41 & 78.00  & 9.71 & 30.80 & 49.52 \\ 
&PKD curriculum     & 35.10 & 48.67 & 59.20 & 47.17 & 62.23 & 75.34 & 49.11 & 60.72 & 79.98  & 10.10 & 31.25 & 50.20\\ 
&PKD anti-curriculum & 34.61 & 47.93 & 58.39 & 46.44 & 61.43 & 74.67 & 48.70 & 60.02 & 79.37  & 8.80 & 26.38 & 44.54\\ \hline

\hline \hline 
\multirow{3}{*}{HMDB-51 } &I-frame KD               & 19.65 & 25.95 & 35.38 & 23.98 & 33.74 & 44.71 & 27.05 & 34.85 & 53.74  & 5.88  & 19.63 & 26.66\\ 
& PKD curriculum     & 20.10 & 27.43 & 38.08 & 24.77 & 35.27 & 48.07 & 27.32 & 35.15 & 53.84  & 5.99 & 19.81 & 26.78\\ 
& PKD anti-curriculum & 19.54 & 26.24 & 36.21 & 24.61 & 35.33 & 47.30 & 26.78 & 34.14 & 52.68  & 5.57  & 16.32 & 23.12\\ \hline
\end{tabular}
\end{adjustbox}
\caption{Accuracy results for IC1, IC2, and IC3 when performing KD with I-frame only (I-frame KD), PKD curriculum, which performs KD between MV, then R and then I-frame (our proposal), and PKD anti-curriculum performs KD between I-frame, then R and then MV (also ours). The reported accuracy is the average over the three testing splits and over three different seeds for UCF-101 and HMDB-51 datasets.}
\label{order_KD}
\vspace{-4mm}
\end{table*}}

\subsection{Analysis of Teacher Classifier Order for PKD}
\label{Analysis of Teacher Classifier Order for PKD}

\begin{wraptable}{l}{0.6\textwidth}
\centering
\begin{adjustbox}{width=0.6\textwidth}
\begin{tabular}{|c|c|cc|cc|}
\hline
\multirow{2}{*}{Method}              &      \multirow{2}{*}{Modalites}   & \multicolumn{2}{c|}{Accuracy}          & \multicolumn{2}{c|}{GFLOPs}                \\ \cline{3-6}
      &                       & \multicolumn{1}{c|}{UCF-101} & HMDB-51 & \multicolumn{1}{c|}{UCF-101}    & HMDB-51 \\ \hline \hline
\rowcolor[HTML]{DAE8FC} 
Liu et al. \cite{liu2023learning} & CD+RD & \multicolumn{1}{c|}{95.8\%} & 73.5\% & \multicolumn{1}{c|}{ $>$ 543,903}    &  $>$ 222,615\\ 
\rowcolor[HTML]{DAE8FC} 
MFCD-Net \cite{battash2020mimic}                                    & CD+RD    & \multicolumn{1}{c|}{93.2\%}  & 66.9\%  & \multicolumn{1}{c|}{1,328,536}    &     543,764    \\ \hline \hline
\rowcolor[HTML]{ECE6DD} 
CoViAR \cite{wu2018compressed}    & CD       & \multicolumn{1}{c|}{93.1\%}  & 68.0\%  & \multicolumn{1}{c|}{543,903 } &  222,615       \\ 
\rowcolor[HTML]{ECE6DD} 
CV-C3D \cite{dos2019cv}                                          & CD       & \multicolumn{1}{c|}{83.9\%}  & 55.7\%  & \multicolumn{1}{c|}{284,555}  &   116,465      \\ 
\rowcolor[HTML]{ECE6DD} 
TEAM-NET \cite{wang2021team}   & CD    & \multicolumn{1}{c|}{93.4\%} & 66.1\% & \multicolumn{1}{c|}{646,582} & 264,639 \\ 
\rowcolor[HTML]{ECE6DD} 
DMC-Net \cite{shou2019dmc}  & CD+OF & \multicolumn{1}{c|}{92.3\%} & 71.8\% & \multicolumn{1}{c|}{31,264,121}  &  12,796,286 \\ \hline \hline
\rowcolor[HTML]{DAD7B3}
Wu et al. \cite{wu2019multi}     & CD       & \multicolumn{1}{c|}{88.5\%}  & 56.2\%  & \multicolumn{1}{c|}{4,717,060} &    1,930,655     \\ 
\rowcolor[HTML]{DAD7B3} 
MIMO \cite{terao2023efficient}         & CD       & \multicolumn{1}{c|}{85.8\%}  & 58.6\%  & \multicolumn{1}{c|}{172,587}  & 70,639        \\ 
\rowcolor[HTML]{DAD7B3}
\textbf{Ours} & CD       & \multicolumn{1}{c|}{\textbf{88.4\%}} &    \textbf{60.3\%}     & \multicolumn{1}{c|}{\textbf{147,391}}  &   \textbf{56,340}      \\ \hline
\end{tabular}
\end{adjustbox}
\caption{SOTA comparison: 1) works using both CD and RD during training (blue), 2) works using only CD but with high computational cost (orange), 3) works focusing on optimizing the computational cost (green). CD: Compressed Domain, RD: Raw Domain, OF: Optical Flow. CoViAR backbone is used for our proposal.}
\label{table_SOTA}
\vspace{-2mm}
\end{wraptable}

\textbf{Experiment.} This analysis considers three approaches. Firstly, we evaluate a structured curriculum, termed PKD curriculum, which orders the FCs of the MV, R, and I-frame backbone networks (teacher network) from the highest to lowest accuracy. PKD curriculum is doing KD using MV, followed by R, and then I-frame FCs as teachers. Secondly, we evaluate the anti-curriculum PKD, starting with I-frame, followed by R, and then MV. Finally, we do KD using only the I-frame final classifier as the teacher. This analysis aims to assess the efficacy of progressive knowledge transfer of the compressed video modalities against the standard approach of KD from a single, highly accurate teacher. To isolate the effect of the training order, we report results \emph{without} using lateral connections. The results reported are the average over the three test splits and three seeds in Table \ref{order_KD} using 3 frames for each compressed video modality.

\noindent \textbf{Results.} From the results in Table \ref{order_KD}, we observe that the PKD curriculum performs the best by up to 2\% and 3\% over I-frame KD on UCF-101 and HMDB-51 respectively. This observation aligns with insights from prior studies in the image classification domain \cite{you2017learning,kaplun2022knowledge}, underscoring that KD from the most accurate classifier may not always ensure the most efficient knowledge transfer.
\vspace{-2mm}
\subsection{Comparison with Prior Works}
\label{Comparison with Prior Works}

\begin{figure}[t]
\centering
\includegraphics[width=0.75\linewidth]{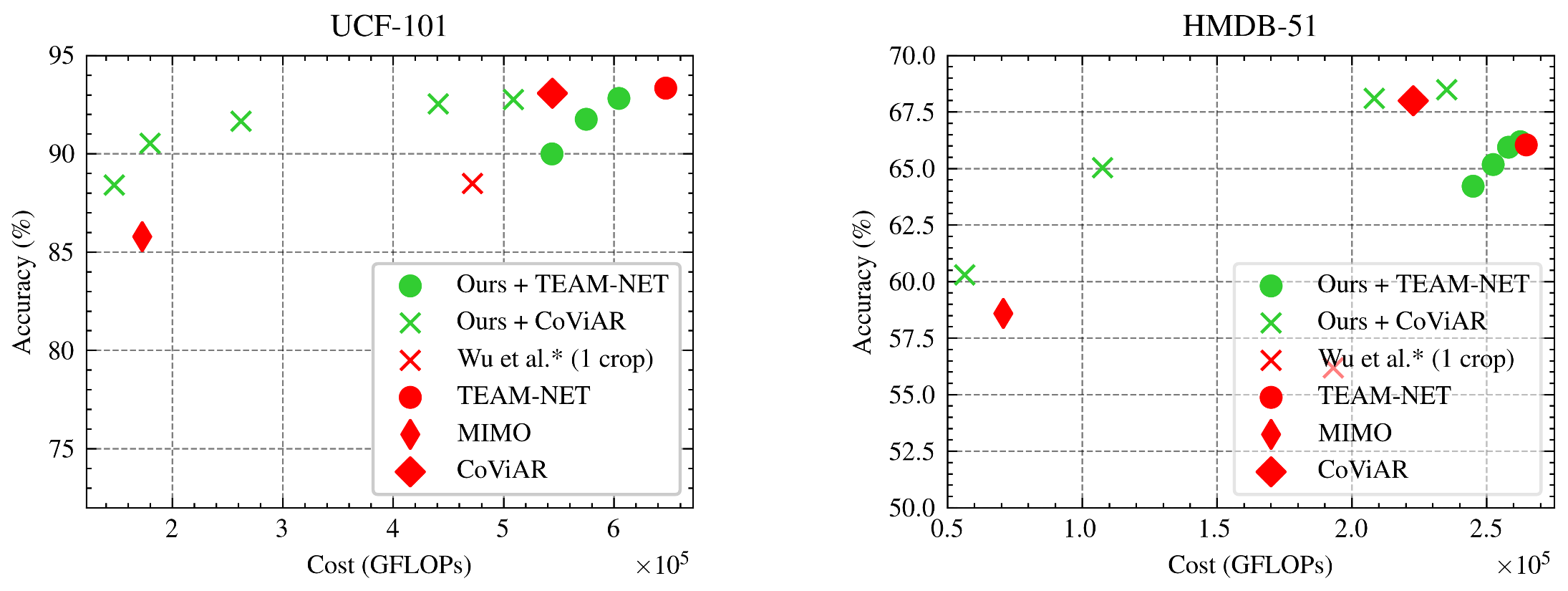} %trade.png
\vspace{-4mm}
\caption{Trade-off curve of our proposal and SOTA work comparison on UCF-101 and HMDB-51. SOTA works (marked in red) fall either in the low accuracy and low computation cost regime (lower-left of the plot) or in the high accuracy and computation cost regime (top-right of the plot). Our proposal uses two backbones, i.e. CoViAR and TEAM-NET, (marked in green) scales in accuracy and computational cost.}
\label{trade_off_figure_HMDB51}
\vspace{-4mm}
\end{figure}

\textbf{Experiment.} We plot the trade-off curve between accuracy and computational cost (Giga Floating point operations - GFLOPs) in Figure \ref{trade_off_figure_HMDB51} for UCF-101 and HMDB-51. We report the cumulative GFLOPs for all three testing splits UCF-101 and HMDB-51. We compare with works that optimize for computational cost. Moreover, Table \ref{table_SOTA} summarizes the state-of-the-art (SOTA) methods: works using both Compressed Domain (CD) and Raw Domain (RD) during training (highlighted in blue), 2) works using only CD but with high computational cost (highlighted in orange), 3) works focusing on optimizing the computational cost (highlighted in green). Our proposal falls in the third category, and we compare it with these works in the trade-off plots (Figure \ref{trade_off_figure_HMDB51}).

\noindent \textbf{Results.} Figure \ref{trade_off_figure_HMDB51} plots the accuracy-compute cost trade-off for UCF-101 and HMDB-51 of our proposal and SOTA methods. The trade-off is achieved by changing the threshold value $\tau$ of WISE. Note, that the threshold value is the same across all the ICs. We observe that MIMO \cite{terao2023efficient} has a very low computational cost but trades off accuracy, while CoViAR \cite{wu2018compressed} is at the other end of the spectrum where it trades off compute cost for accuracy. Our approach with PKD and WISE can efficiently scale between these two extremes by changing the threshold. Notably, our proposal showcases a reduction in GFLOPs under the iso-accuracy scenario, surpassing the efficiency of CoViAR \cite{wu2018compressed}, and the method proposed by Wu et al.  \cite{wu2019multi}. On the other extreme when compared to MIMO \cite{terao2023efficient} it achieves better accuracy at a similar compute cost for UCF-101 and significantly improved accuracy at the lower compute cost for the HMDB-51 dataset. Further, compared with MIMO \cite{terao2023efficient} work, our proposal has better latency and bandwidth efficiency, as reported in the Appendix.

\section{Conclusions}
\label{Conclusions}

PKD uses the hierarchical nature of neural network convergence across the different video modalities: motion vectors, residuals, and intra-frames and facilitates the sequential transfer of knowledge leading to better generalization. The results of our experiments highlight the superiority of PKD over CE based training methods. Additionally, WISE further enhances inference accuracy by optimally combining outputs from the ICs. These findings underscore the potential of our proposed techniques to significantly advance the field of video action recognition, offering a promising avenue for future research and application.

\section*{Acknowledgments}

This work was supported in part by, the Center for the Co-Design of Cognitive Systems (CoCoSys), a DARPA-sponsored JUMP 2.0 center, the Semiconductor Research Corporation (SRC), and the National Science Foundation.

\bibliography{egbib}

% \documentclass{bmvc2k}

% % \usepackage[table,xcdraw]{xcolor}
% \usepackage{multirow}
% \usepackage{adjustbox}
% \usepackage{colortbl}

% \usepackage{graphicx} % Required for including images
% \usepackage{subcaption} % Required for creating subfigures
% \usepackage{wrapfig}

%% Enter your paper number here for the review copy
% \bmvcreviewcopy{948}

\section*{Appendix}
% Enter the paper's authors in order
% \addauthor{Name}{email/homepage}{INSTITUTION_CODE}
% \addauthor{Susan Student}{http://www.vision.inst.ac.uk/~ss}{1}
% \addauthor{Petra Prof}{http://www.vision.inst.ac.uk/~pp}{1}
% \addauthor{Colin Collaborator}{colin@collaborators.com}{2}

% % Enter the institutions
% % \addinstitution{Name\\Address}
% \addinstitution{
%  The Vision Institute\\
%  University of Borsetshire\\
%  Wimbleham, UK
% }
% \addinstitution{
%  Collaborators, Inc.\\
%  123 Park Avenue,\\
%  New York, USA
% }

% \runninghead{Student, Prof, Collaborator}{BMVC Author Guidelines}
% \begin{document}

% \maketitle

The appendix reports the training and inference hyperparameters in Section \ref{Training and Inference Hyperparameters}. Section \ref{Loss Landscape Visualization} presents the loss landscape results for Cross-Entropy (CE) and Progressive Knowledge Distillation (PKD) for the ICs for the compressed video modalities networks. Our method's latency and bandwidth results are summarized in Section \ref{Latency and Bandwidth Results}. Section \ref{sec:standard_deviation_tables} reports the standard deviation of the results in the main paper. The experiments were run for the three testing splits over three different seeds (i.e., 3 seeds for each split, thus, 9 data points in total for one entry). Section \ref{sec:study_frames} analyses the effect of the number of frames for each compressed video modality on the accuracy and the computational cost. Note that the supplementary material includes the code for reproducing our experiments (the README.md file contains the details on how to run the scripts for the experiments).

\subsection*{Training and Inference Hyperparameters}
\label{Training and Inference Hyperparameters}

\textbf{Training.} All videos were resized to 240 × 320 resolution and compressed to the MPEG4 Part-2 format \cite{le1991mpeg}. We randomly sampled 3 Intra-Frames (I-frames), Motion Vectors (MVs), and Residuals (Rs) from videos allocated for the training. The networks for the UCF-101 and HMDB-51 were initialized with pre-trained models on the Kinetics dataset \cite{kay2017kinetics}. Further optimization was carried out on UCF-101 and HMDB-51, employing mini-batch training and the Adam optimizer \cite{kingma2014adam} with a weight decay of 0.0001, an epsilon value of 0.001, an initial learning rate of 0.003 for the I-frame input, a learning rate of 0.01 for the MV input, and a learning rate of 0.005 for the R input. The backbone networks were trained for 510 epochs, with a decay in learning rate by a factor of 0.1 at the 150th, 270th, and 390th epochs.

ICs were attached after each residual block of the backbone networks. The IC architecture consists of one convolutional layer and one fully connected layer. The ICs were optimized for 150 epochs using the Adam optimizer, starting with the same initial learning rate as the backbone networks. The learning rate for the ICs underwent decay by a factor of 0.1 at the 50th, 100th, and 150th epochs. The temperature value for the KD was set to 1. The training and evaluation were conducted on a server equipped with an Intel(R) Xeon(R) Silver 4114 CPU and 4 NVIDIA GeForce GTX 1080Ti GPUs with 12 GB of video memory.

\noindent \textbf{Inference.} During inference, we use 1 I-frame, 1 MV, and 2 R segments when using the CoViAR backbone, and 8 I-frames, 8 MV, and 8 R segments for the TEAM-Net backbone unless stated otherwise. The frames were cropped into 224 × 224 patches and underwent horizontal flipping with 50\% probability (i.e., we used only 1 crop). The entire framework was implemented using PyTorch \cite{paszke2019pytorch}.

\subsection*{Loss Landscape Visualization}
\label{Loss Landscape Visualization}

\begin{figure}
\includegraphics[width=0.9\linewidth]{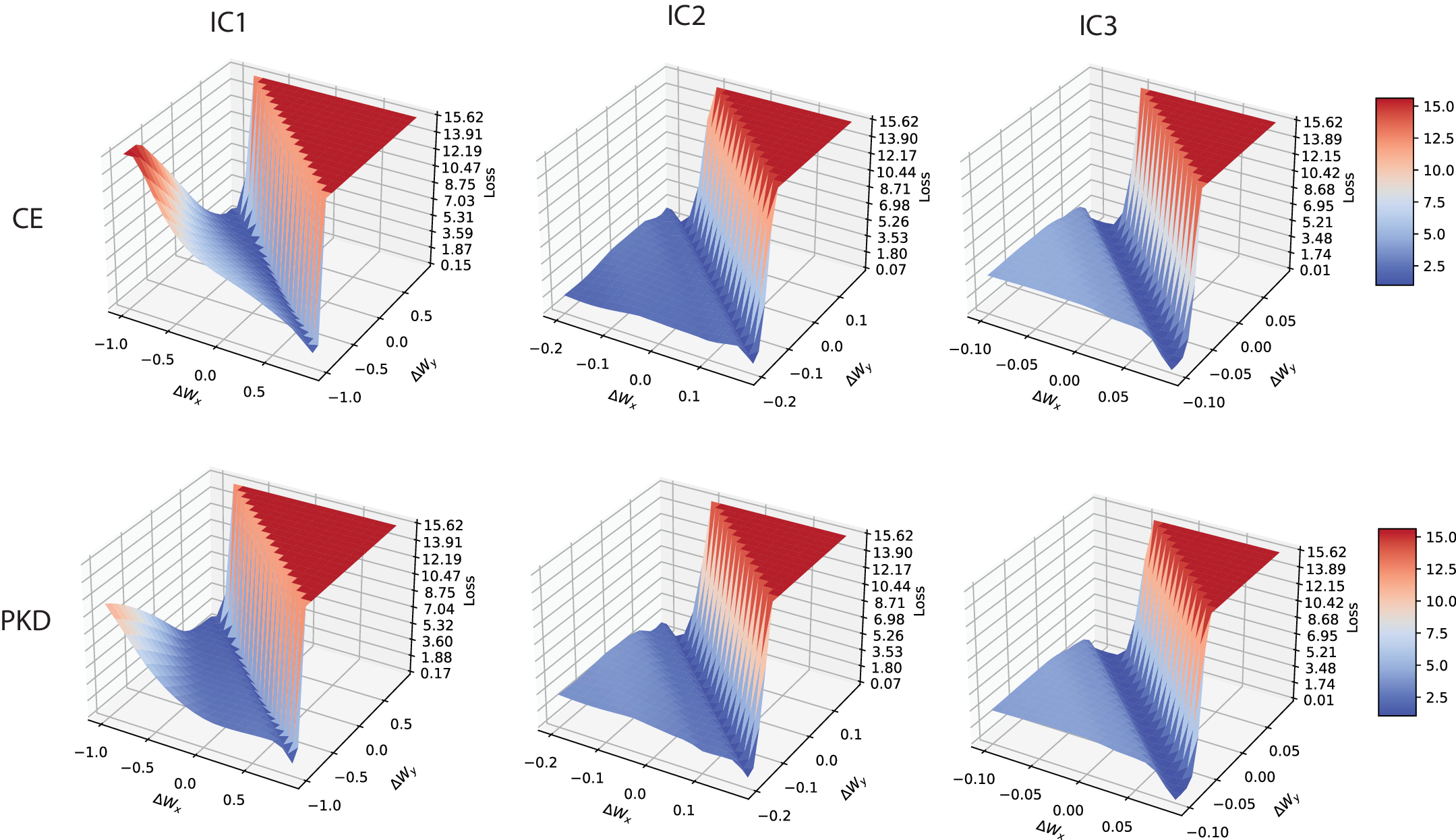}
\caption{Visualizing the loss surface around the minimum of early exit classifiers attached to the MV backbone network when trained using CE vs PKD.}
\label{mv_vis}
\end{figure}

\begin{figure}
\includegraphics[width=0.9\linewidth]{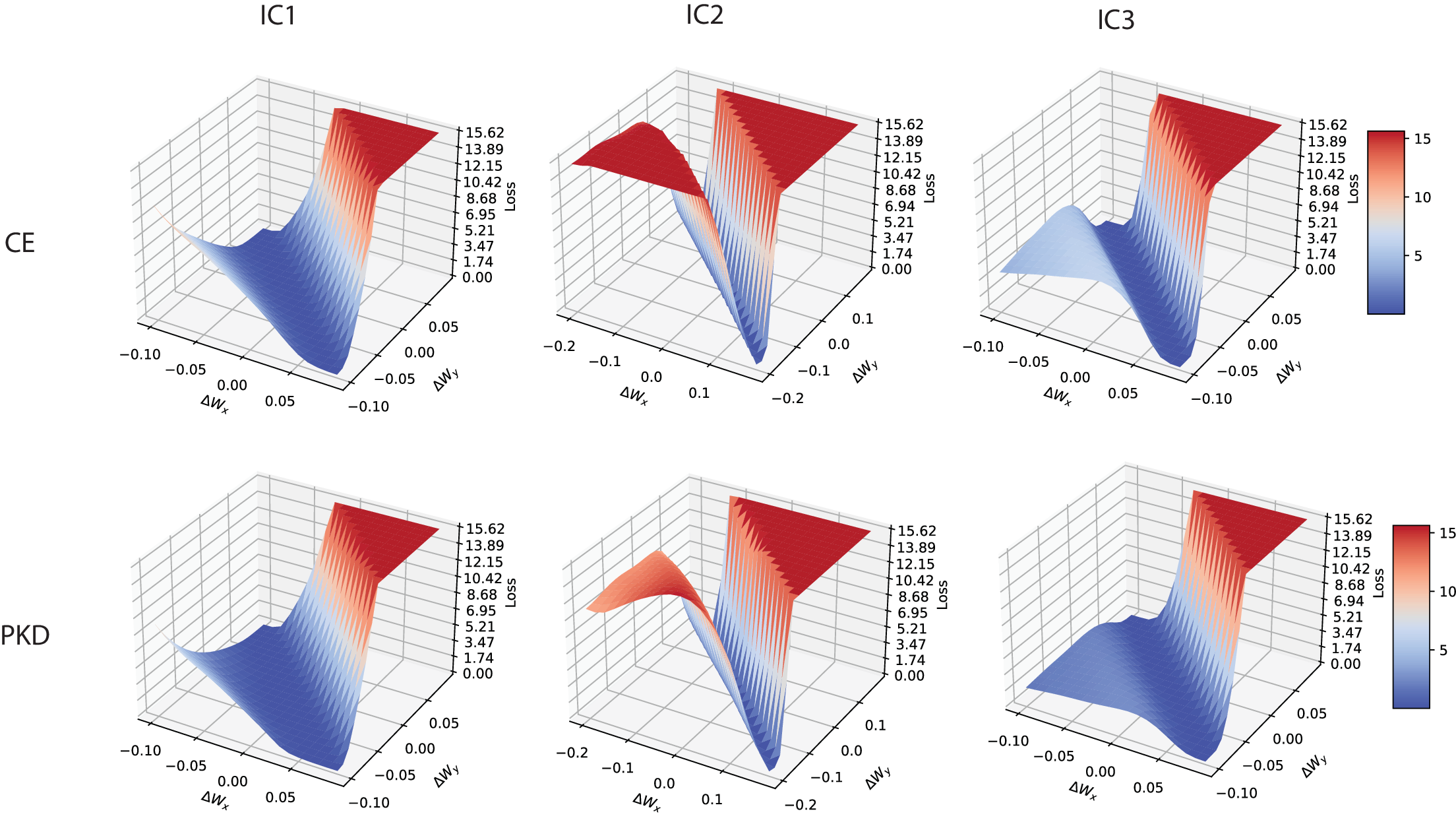}
\caption{Visualizing the loss surface around the minimum of early exit classifiers attached to the Residual backbone network when trained using CE vs PKD.}
\label{residual_vis}
\end{figure}

\begin{figure}
\includegraphics[width=0.9\linewidth]{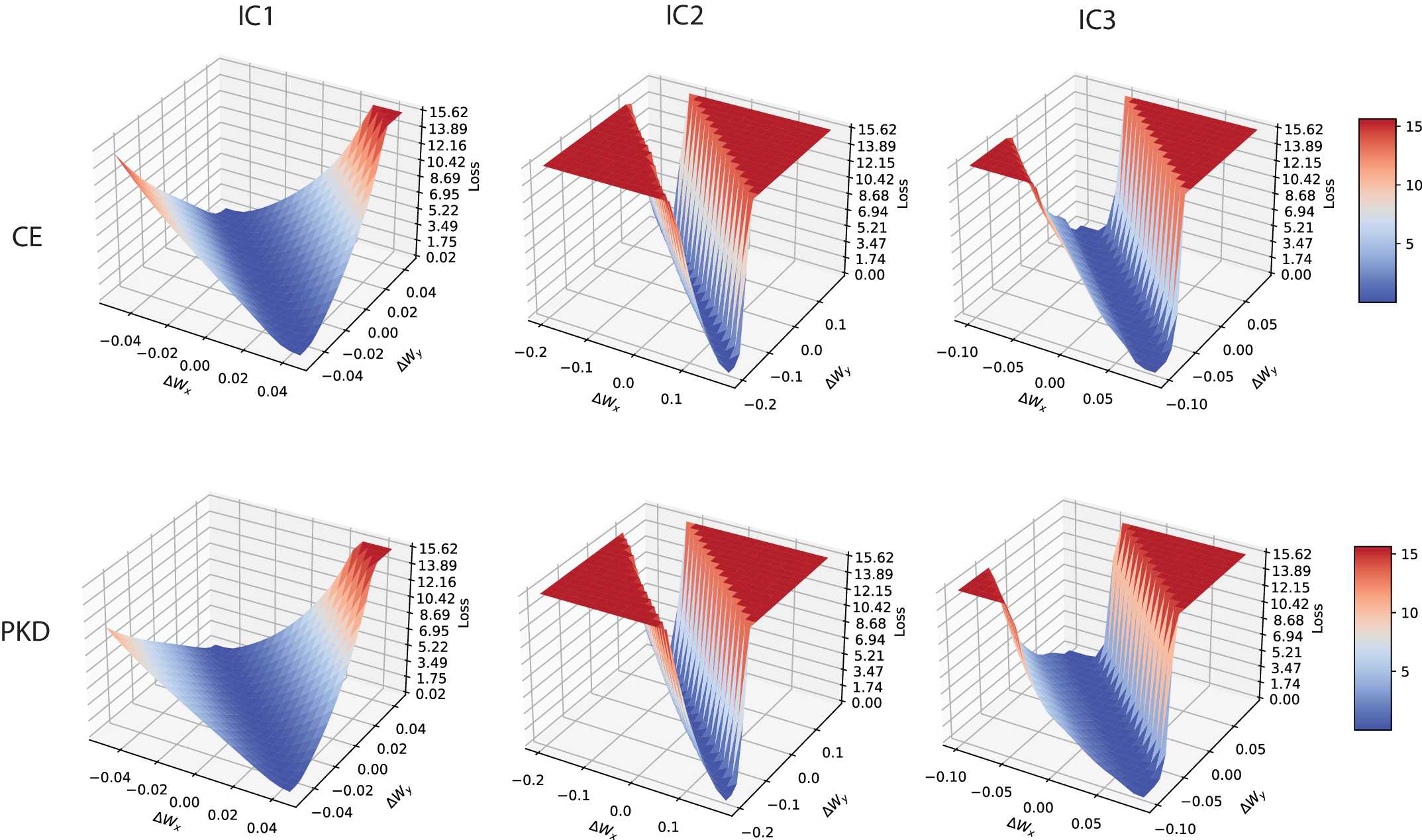}
\caption{Visualizing the loss surface around the minimum of early exit classifiers attached to the I-frame backbone network when trained using CE vs PKD.}
\label{iframe_vis}
\end{figure}

Figures \ref{mv_vis},  \ref{residual_vis}, \ref{iframe_vis} visualize the loss landscape for the ICs of the Motion Vector, Residual, and I-frame network, respectively, trained with CE and PKD. We observe that PKD converges to flatter minimum compared to CE. Flatter minima are generally associated with better generalization \cite{hochreiter1994simplifying, hochreiter1997flat}.

\subsection*{Latency and Bandwidth Results}
\label{Latency and Bandwidth Results}

We evaluate our proposal using the latency and bandwidth metrics. Consider the following scenario: video is transmitted in the form of packets that include MV, R, and I-frames. To classify a video stream, it is necessary to receive a sufficient number of these packets, as required by each classification technique. Therefore, we assess the latency based on the time interval required to classify a video stream at a consistent bit rate (i.e., iso-bandwidth). Our results provide relative latency between different SOTA methods for convenience. 

To compare bandwidth effectively, we align the latency across different methods. Specifically, we determine how much bandwidth other techniques would require to match the latency of our method. This approach allows us to measure and compare the bandwidth efficiency of various techniques, ensuring they all meet the same latency target (i.e., iso-latency). 

We use the CoViAR backbone with the proposed PKD methodology for this comparison. CoViAR backbone uses a ResNet18 for MV, a ResNet18 for R, and a ResNet50 for I-frames. This allows using different numbers of frames for MV, R, and I-frames. For example, we can use 3 MV, 2 R, and 5 I-frames. In comparison, MIMO \cite{terao2023efficient} uses a single network architecture, and this imposes the constraint of using the same number of MV, R, and I-frames. The authors of MIMO \cite{terao2023efficient} report results when using 8 MV, 8 R, and 8 I-frames in their experiments, and we follow the same when comparing with MIMO. The comparison results of our method with SOTA works for latency and bandwidth metrics are reported in Table \ref{table_latency_bandwidth}. Notably, our proposal is 3x and 8x times more efficient than Wu et al. and MIMO in terms of latency and 3.3x and 2.09x more efficient than Wu et al. and MIMO in terms of bandwidth. Please note that, unlike the CoViAR backbone, the TEAM-Net backbone requires the same number of MV, R, and I-frames as input and thus does not benefit from the proposed PKD methodology.

\begin{table}[t]
\centering
\begin{tabular}{|c|c|c|}
\hline 
Method & Latency & Bandwidth \\ \hline \hline
MIMO \cite{terao2023efficient} & 8x & 2.09x \\ 
Wu et al. \cite{wu2019multi} & 3x & 3.3x \\ 
CoViAR \cite{wu2018compressed} & 3x & 3.3x \\ 
Our Method & 1x & 1x \\ \hline
\end{tabular}
\caption{Latency and Bandwidth comparison results.}
\label{table_latency_bandwidth}
\end{table}

\subsection*{Tables with Standard Deviation Values}
\label{sec:standard_deviation_tables}

In this section, we summarize the results of the tables from the main paper and report the mean and standard deviation values of the 3 splits across 3 random seeds. Please note that in the main paper, we reported only the mean values for compactness. Table \ref{teamnet_progkd} summarizes the results for PKD for TEAM-Net backbone on UCF-101 and HMDB-51 datasets, and Table \ref{table_combined_results} summarizes the results for PKD for CoViAR backbone on UCF-101 and HMDB-51 datasets. Table \ref{table_wise} summarizes the results (mean values and standard deviations) for WISE applied on the CoViAR and TEAM-Net backbones on UCF-101 and HMDB-51 datasets.

% {\renewcommand*{\arraystretch}{1.3}
\begin{table}[t]
\centering
\begin{adjustbox}{width=0.8\columnwidth}
\begin{tabular}{|c|c|ccc|} \hline
 &  & \multicolumn{3}{c|}{TEAM-Net Backbone} \\ \cline{3-5} 
Dataset                 & Method     & IC1     & IC2     & IC3     \\ \hline \hline
\multirow{4}{*}{UCF-101} & CE         &  8.86 $\pm$ 1.95 & 28.93 $\pm$ 1.76 & 47.41 $\pm$ 0.92\\
& I-frame KD   & 9.71 $\pm$ 1.95 & 30.80 $\pm$ 1.46 & 49.52 $\pm$ 1.01\\
                        & PKD Curriculum   &  10.10 $\pm$ 1.92 & 31.25 $\pm$ 1.67 & 50.20 $\pm$ 1.09 \\
                        & PKD anti-curriculum    & 8.80 $\pm$ 1.33& 26.38 $\pm$ 1.24& 44.54 $\pm$ 0.87\\
                        \hline \hline 
\multirow{4}{*}{HMDB-51} & CE         & 5.38 $\pm$ 0.70 & 18.12 $\pm$ 1.19 & 24.40 $\pm$ 1.66\\
& I-frame KD    & 5.88 $\pm$ 1.00 & 19.63 $\pm$ 1.28& 26.66 $\pm$ 1.50\\
                        & PKD Curriculum    & 5.71 $\pm$ 0.99 & 19.21 $\pm$ 1.16 & 26.04 $\pm$ 1.29 \\
                        & PKD anti-curriculum    & 5.57 $\pm$ 0.81 & 16.32 $\pm$ 0.91& 23.12 $\pm$ 1.35\\
                         \hline
\end{tabular}
\end{adjustbox}
\caption{Accuracy results for IC1, IC2, and IC3 when using Cross Entropy (CE), performing KD with I-frame only (I-frame KD), PKD curriculum, which performs KD between MV, then R and then I-frame (our proposal), and PKD anti-curriculum performs KD between I-frame, then R and then MV (also ours). The reported accuracy is the average over \emph{three testing splits} and \emph{three different seeds} (9 datapoints per entry) for UCF-101 and HMDB-51 datasets using TEAM-Net backbone.}
\label{teamnet_progkd}
\end{table}
% }

% {\renewcommand*{\arraystretch}{1.2}
\begin{table*}[t]
\centering
\begin{adjustbox}{width=\textwidth}
\begin{tabular}{|c|c|ccc|ccc|ccc|}
\hline
\multirow{3}{*}{Dataset} & \multirow{3}{*}{Method} & \multicolumn{9}{c|}{CoViAR Backbone} \\ \cline{3-11} 
& &  \multicolumn{3}{c|}{MV} & \multicolumn{3}{c|}{Residual}  & \multicolumn{3}{c|}{I-Frame} \\ 
& & IC1 & IC2 & IC3 & IC1 & IC2 & IC3 & IC1 & IC2 & IC3  \\ \hline \hline
\multirow{3}{*}{UCF-101} & CE & 34.39  $\pm$  0.41 & 47.29  $\pm$  0.90 & 57.43  $\pm$  1.70 & 46.45  $\pm$  0.79 & 60.57  $\pm$  0.72 & 72.63  $\pm$  0.61 & 48.40  $\pm$  0.44 & 57.61  $\pm$  1.05 & 74.11  $\pm$  1.12\\
& I-frame KD               & 34.51 $\pm$ 0.42 & 47.19 $\pm$ 0.85 & 57.38 $\pm$  1.85 & 46.70 $\pm$ 0.48 & 60.92 $\pm$ 0.40 & 73.25  $\pm$ 0.34 & 48.93 $\pm$ 0.36 & 59.41 $\pm$ 1.08 & 78.00  $\pm$  0.83 \\ 
& PKD curriculum & 35.10  $\pm$  0.45 & 48.67  $\pm$  1.24 & 59.20  $\pm$  1.72 & 47.17  $\pm$  0.67 & 62.23  $\pm$  0.73 & 75.34  $\pm$  0.21 & 49.11  $\pm$  0.66 & 60.72  $\pm$  1.83 & 79.98  $\pm$  1.82 \\
& PKD anti-curriculum & 34.61 $\pm$ 0.58 & 47.93 $\pm$ 0.93 & 58.39  $\pm$ 1.74 & 46.44 $\pm$ 0.75 & 61.43 $\pm$ 0.66 & 74.67  $\pm$  0.48 & 48.70 $\pm$ 0.55 & 60.02 $\pm$ 0.86 & 79.37  $\pm$  0.51 \\ \hline \hline

\multirow{3}{*}{HMDB-51} & CE & 19.92  $\pm$  0.83 & 26.24  $\pm$  0.66 & 34.58  $\pm$  1.03 & 21.78  $\pm$  0.64 & 31.13  $\pm$  0.85 & 40.19  $\pm$  0.42 & 23.55  $\pm$  1.04 & 27.09  $\pm$  0.86 & 42.42  $\pm$  1.14 \\
& I-frame KD               & 19.65 $\pm$ 0.90 & 25.95 $\pm$ 0.61 & 35.38  $\pm$  0.95 & 23.98 $\pm$ 0.90 & 33.74 $\pm$ 0.68 & 44.71  $\pm$  1.06 & 27.05 $\pm$ 0.50 & 34.85 $\pm$ 1.32 & 53.74  $\pm$  0.67 \\ 
& PKD curriculum & 20.10  $\pm$  0.96 & 27.43  $\pm$  0.70 & 38.08  $\pm$  0.52 & 24.77  $\pm$  0.94 & 35.27  $\pm$  0.71 & 48.07  $\pm$  0.38 & 27.32  $\pm$  0.63 & 35.15  $\pm$  1.16 & 53.84  $\pm$  1.16 \\ 
& PKD anti-curriculum & 19.54 $\pm$ 1.15 & 26.24 $\pm$ 0.90 & 36.21  $\pm$  0.87 & 24.61 $\pm$ 1.16 & 35.33 $\pm$ 0.83 & 47.30  $\pm$  0.45 & 26.78 $\pm$ 0.27 & 34.14 $\pm$ 1.61 & 52.68  $\pm$  0.67 \\ \hline
\end{tabular}
\end{adjustbox}
\caption{Accuracy results for IC1, IC2, and IC3 when using Cross Entropy (CE), performing KD with I-frame only (I-frame KD), PKD curriculum, which performs KD between MV, then R and then I-frame (our proposal), and PKD anti-curriculum performs KD between I-frame, then R and then MV (also ours). The reported accuracy is the average over \emph{three testing splits} and \emph{three different seeds} (9 datapoints per entry) for UCF-101 and HMDB-51 datasets using CoViAR backbone.}
\label{table_combined_results}
\end{table*}
% }

% {\renewcommand*{\arraystretch}{1.2}
\begin{table}[hbt!]
\centering
\begin{adjustbox}{width=0.8\columnwidth}
\begin{tabular}{|c|c|cc|}
\hline
 \multirow{2}{*}{Backbone}                      & \multirow{2}{*}{Method}                      & \multicolumn{2}{c|}{Accuracy}          \\ \cline{3-4}
 &                      & \multicolumn{1}{c|}{UCF-101} & HMDB-51 \\ \hline \hline
\multirow{3}{*}{CoViAR} & No Lateral Connections  & \multicolumn{1}{c|}{88.59\% $\pm$ 0.18 }   & 58.91\%   $\pm$  0.93 \\ 
& Uniform Scaling  & \multicolumn{1}{c|}{92.08\% $\pm$ 0.03}   & 67.50\%  $\pm$  0.40 \\ 
& WISE (Ours)             & \multicolumn{1}{c|}{\textbf{92.87\% $\pm$ 0.04}}   & \textbf{68.21\% $\pm$ 0.50}  \\ \hline \hline

\multirow{3}{*}{TEAM-Net} & No Lateral Connections  & \multicolumn{1}{c|}{93.24\%  $\pm$  0.15}   & 65.94\%   $\pm$  0.58\\ 
& Uniform Scaling  & \multicolumn{1}{c|}{89.49\%  $\pm$  0.71}   & 61.89\%  $\pm$  0.68\\ 
& WISE (Ours)             & \multicolumn{1}{c|}{\textbf{93.29\% $\pm$ 0.52}}   & \textbf{66.29\%  $\pm$  0.23}  \\ \hline
\end{tabular}
\end{adjustbox}
\caption{Results when using: 1) no lateral connections between the ICs, 2) uniform scaling factors and lateral connections between the ICs, 3) our proposed WISE. CoViAR and TEAM-Net are used as the backbone networks. The reported accuracy is the average over the \emph{3 testing splits} and \emph{3 different seeds} (9 datapoints per entry) for UCF-101 and HMDB-51.}
\label{table_wise}
\end{table}
% }

\subsection*{Effect of the Number of MV, R, and I-Frames}
\label{sec:study_frames}
\begin{figure}[t]
    \centering
    \begin{minipage}[t]{0.3\textwidth}
        \centering
        \includegraphics[width=\linewidth]{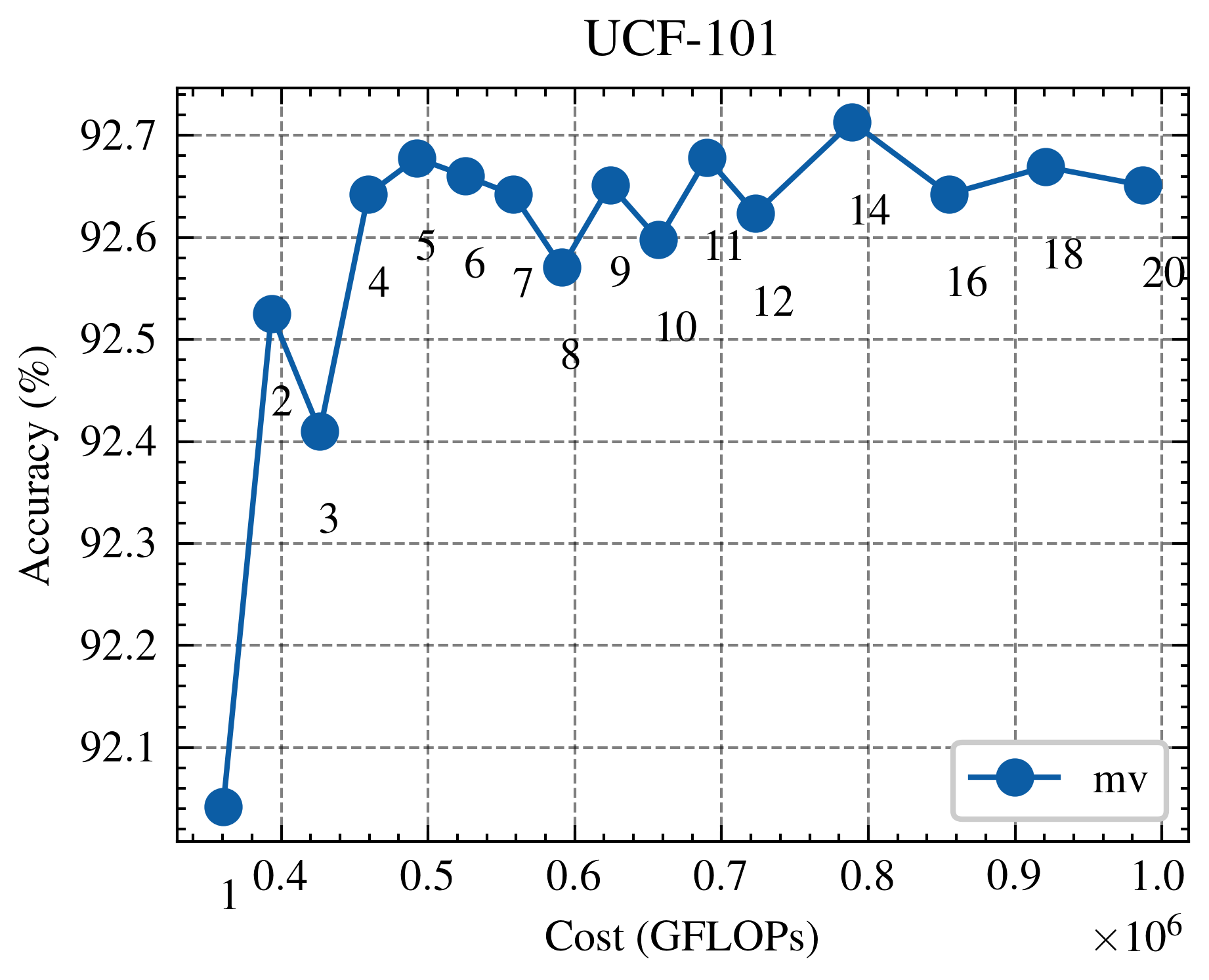}
        \vspace{-2mm}
    \end{minipage}\hfill
    \begin{minipage}[t]{0.3\textwidth}
        \centering
        \includegraphics[width=\linewidth]{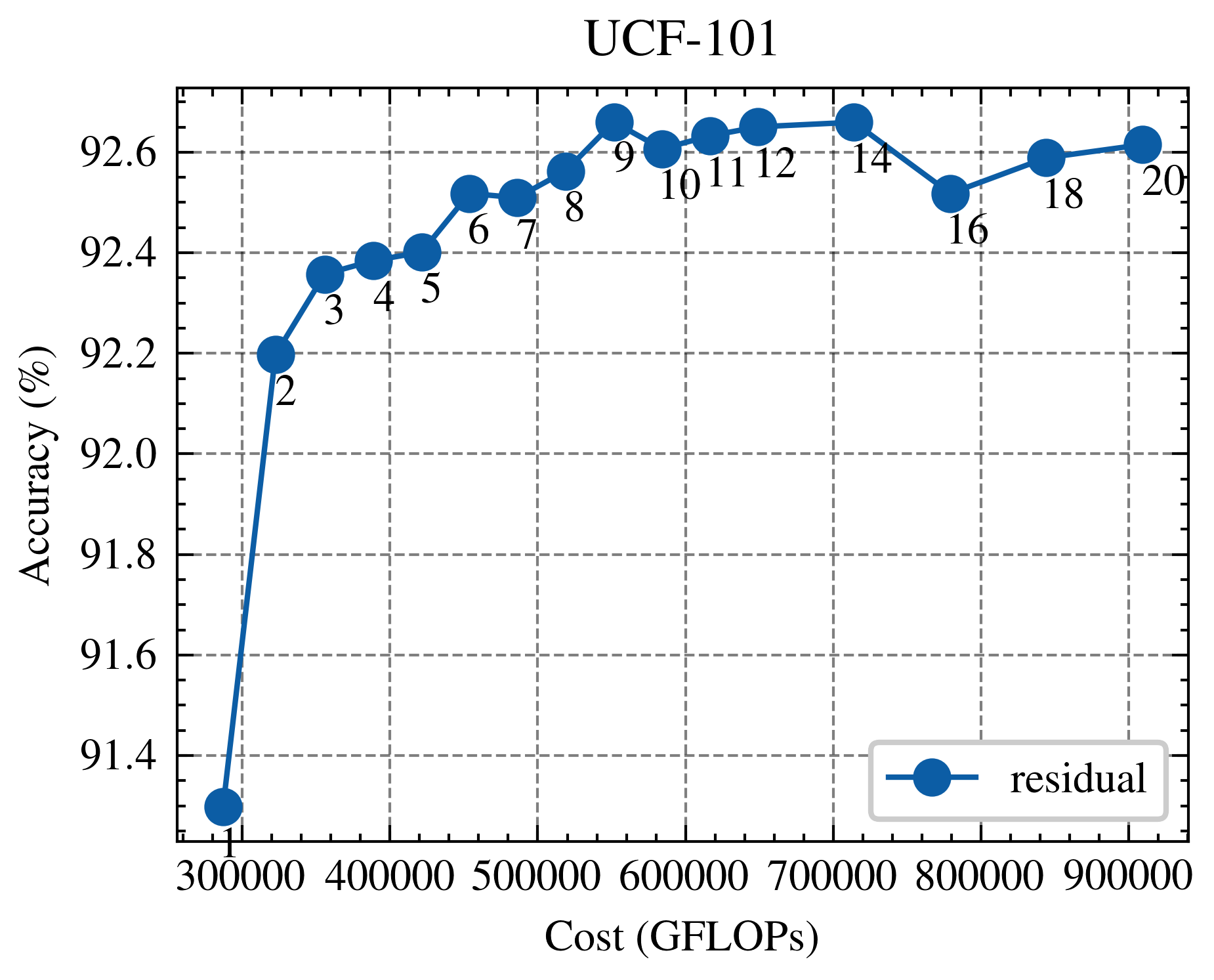}
        \vspace{-2mm}
    \end{minipage}\hfill
    \begin{minipage}[t]{0.3\textwidth}
        \centering
        \includegraphics[width=\linewidth]{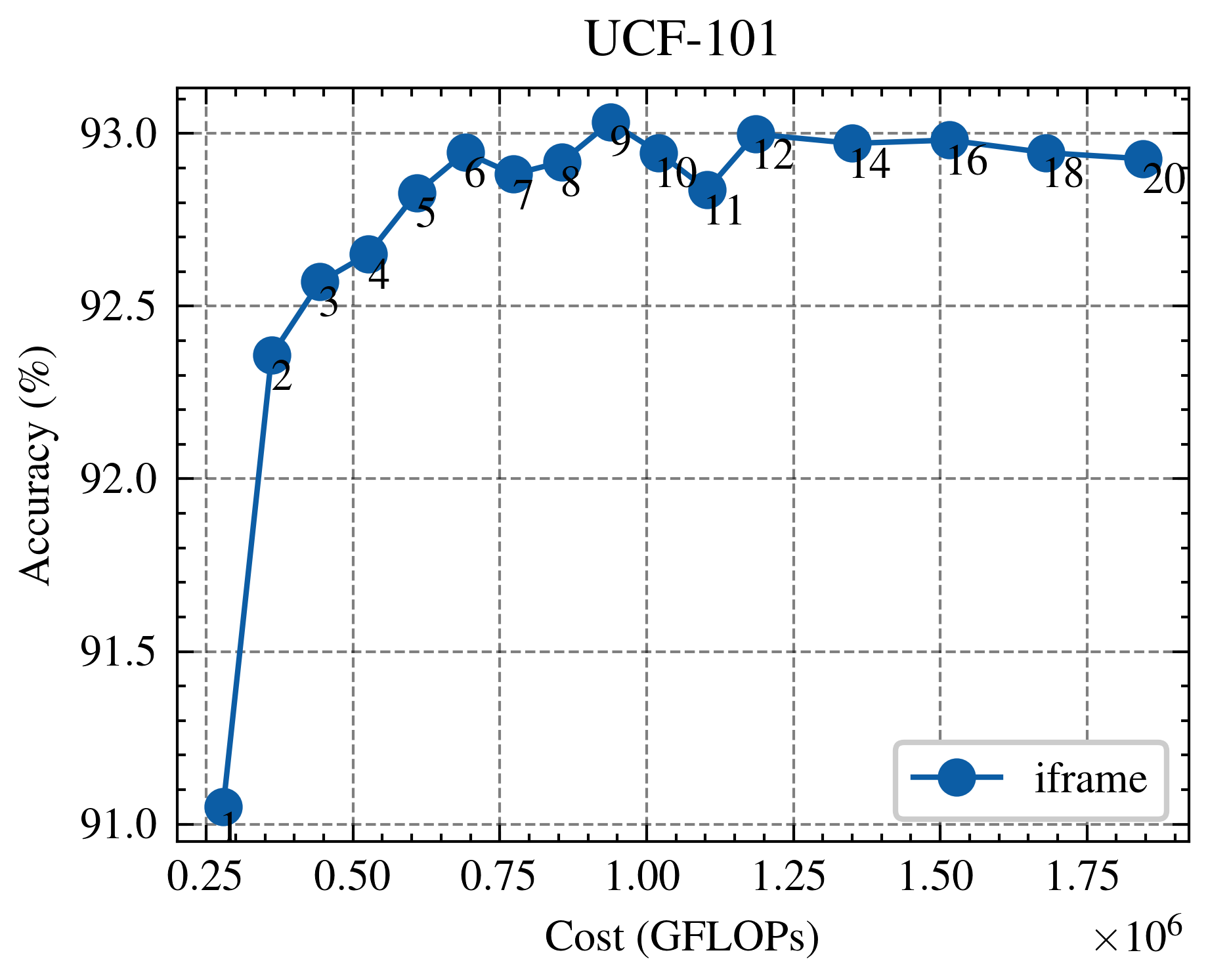}
        \vspace{-2mm}
    \end{minipage}
    \vspace{-4mm}
        \begin{minipage}[t]{0.3\textwidth}
        \centering
        \includegraphics[width=\linewidth]{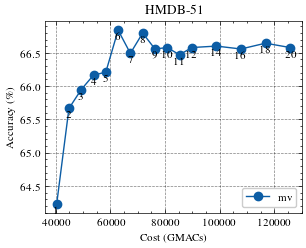}
        \vspace{-2mm}
    \end{minipage}\hfill
    \begin{minipage}[t]{0.3\textwidth}
        \centering
        \includegraphics[width=\linewidth]{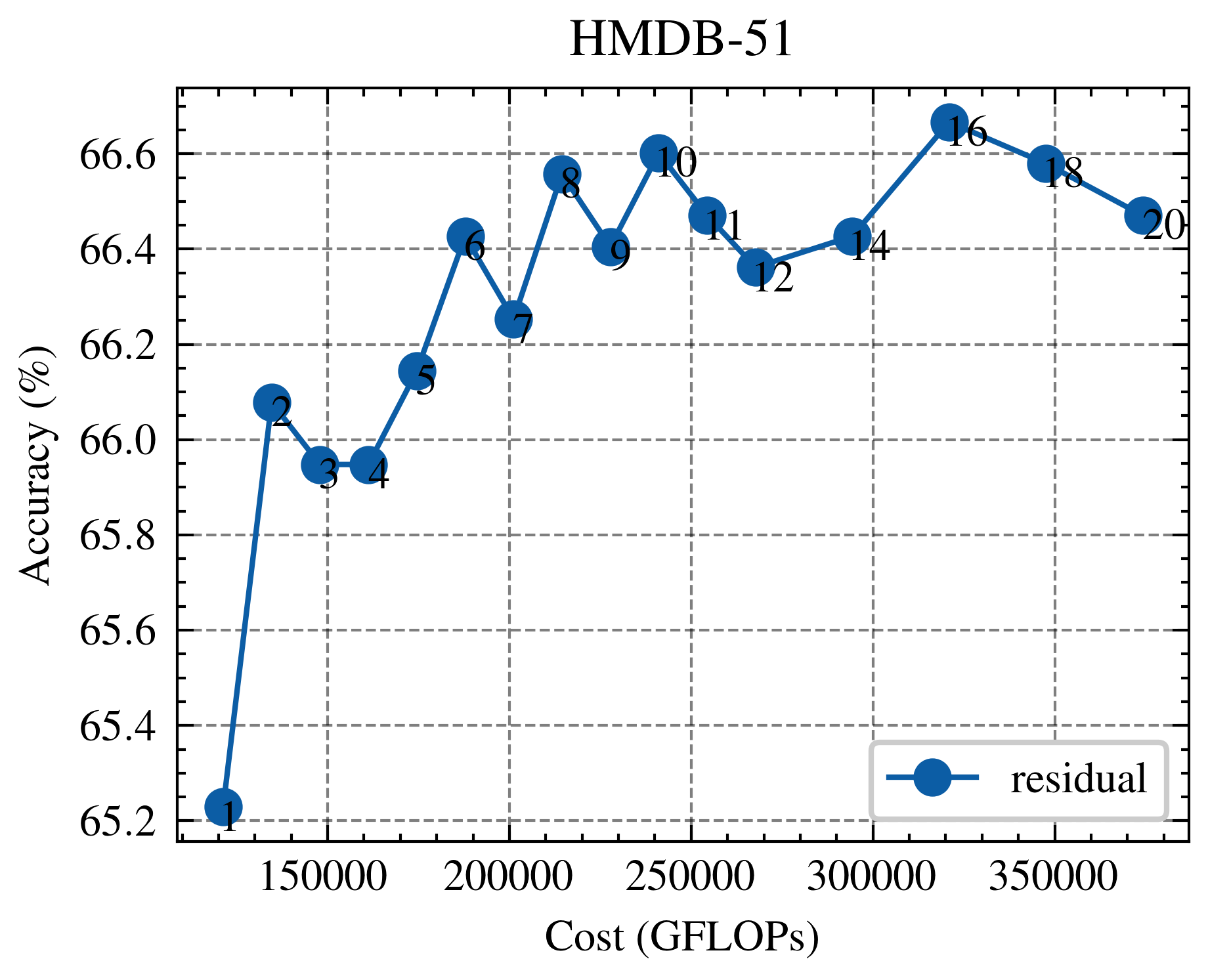}
        \vspace{-2mm}
    \end{minipage}\hfill
    \begin{minipage}[t]{0.3\textwidth}
        \centering
        \includegraphics[width=\linewidth]{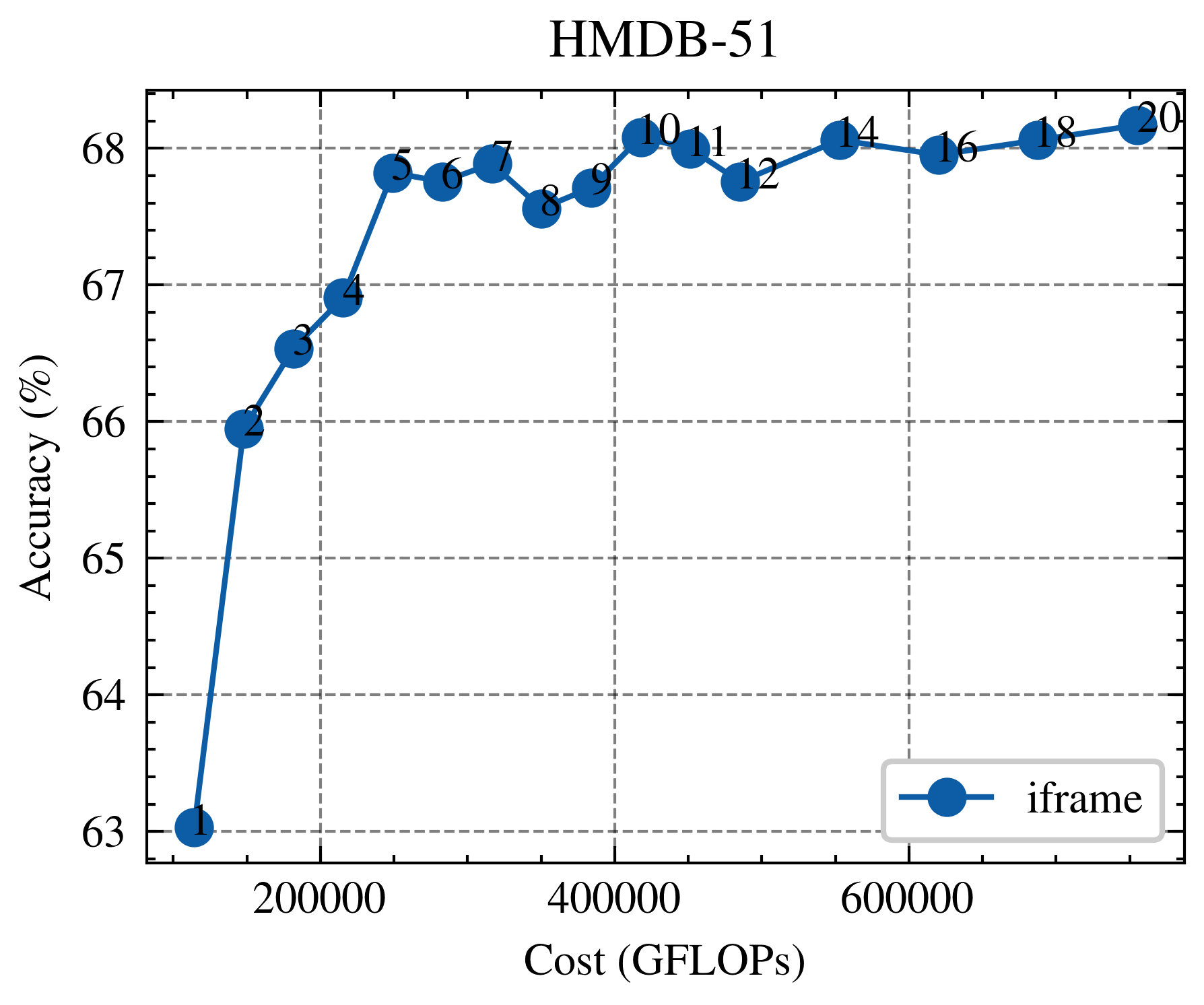}
        \vspace{-2mm}
    \end{minipage}
    \caption{Study on the effect of number of frames when using our proposal with the CoViAR backbone when: 1) increasing the MV frames while keeping the number of R and I-frames constant, 2) increasing the R frames while keeping the number of MV and I-frames constant, and 3) increasing the I-frames while keeping the number of MV and R constant. The first row illustrates the plots for UCF-101, and the second row illustrates the plots for HMDB-51.}
\label{ablation_study_number_of_frames}
    \vspace{-4mm}
\end{figure}

CoViAR \cite{wu2018compressed} backbone employs a ResNet18 for MV, a ResNet18 for R, and a ResNet50 for I-frames. Thus, we explore the impact of using a different number of frames for each compressed video modality (MV, R, I-frame) on the accuracy and the computational cost (Giga Floating point operations - GFLOPs) during inference. TEAM-Net backbone requires the same number of MV, R, and I-frames as input. Therefore, this analysis cannot be done using our proposal with the TEAM-Net backbone.

\noindent \textbf{Experiment.} We perform the following experiments to isolate the effect of utilizing more frames on each compressed video modality:

\begin{enumerate}
\item We increase the number of MV frames while maintaining constant R and I-frames and observe how test accuracy changes (we use 3 R and 3 I-frames, and the threshold is set to 99.99\%).
\item Similarly, we increase R frames while maintaining constant MV and I-frames (we use 3 MV and 3 I-frames, and the threshold is set to 99.99\%).
\item Similarly, we increase I-frames while maintaining constant MV and R frames (we use 3 MV and 3 R frames, and the threshold is set to 99.99\%).
\end{enumerate}

The accuracy was measured using the backbone networks (without the ICs) and the results were averaged over the three test splits. To calculate the GFLOP operations, we used the ptflops library \cite{ptflops}.

\noindent \textbf{Results.}
Figure \ref{ablation_study_number_of_frames} illustrates the effect of using more frames for each compressed video modality. Each plot in Figure \ref{ablation_study_number_of_frames} plots the computational cost (GFLOPs) on the x-axis and accuracy in percentage on the y-axis. Interestingly, we observe a saturation point in accuracy beyond a certain number of frames despite increasing the GFLOPs. This underscores that an excessive number of frames fails to yield accuracy benefits.

% \bibliography{egbib}

% \end{document}

\end{document}